\documentclass{article}

\usepackage[final,nonatbib]{neurips_2019}
\usepackage[utf8]{inputenc} 
\usepackage[T1]{fontenc}    
\usepackage{hyperref}       
\usepackage{url}            
\usepackage{booktabs}       
\usepackage{amsfonts}       
\usepackage{nicefrac}       
\usepackage{microtype}      

\usepackage{graphicx}
\usepackage{authblk}
\usepackage{subfigure}
\usepackage{float}
\usepackage{array}
\usepackage{color}

\title{The Emergence of Compositional Languages for Numeric Concepts Through Iterated Learning in Neural Agents}
\author{%
  Shangmin Guo, Yi Ren, Serhii Havrylov, Stella Frank, Ivan Titov, Kenny Smith\\
  University of Edinburgh \\
  \texttt{s1798190@ed.ac.uk} \\
}

\begin{document}



\maketitle

\begin{abstract}
Since first introduced by \cite{hurford1989biological}, computer simulation has been an increasingly important tool in evolutionary linguistics.
Recently, with the development of deep learning techniques, research in grounded language learning has also started to focus on facilitating the emergence of compositional languages without pre-defined elementary linguistic knowledge.
In this work, we explore the emergence of compositional languages for numeric concepts in multi-agent communication systems.
We demonstrate that compositional language for encoding numeric concepts can emerge through iterated learning in populations of deep neural network agents.
However, language properties greatly depend on the input representations given to agents.
We found that compositional languages only emerge if they require less iterations to be fully learnt than other non-degenerate languages for agents on a given input representation.
\end{abstract}

\section{Introduction}
\label{sec:1.intro}

With recent advances in deep learning (DL), it has been shown that computational agents can master a variety of complex cognitive tasks \cite{mnih2015human, silver2017mastering}.
Recent work in grounded language learning \cite{hermann2017grounded, havrylov2017emergence} applied DL techniques to enable agents to discover through learning communication protocols exhibiting language-like properties, e.g. hierarchy and compositionality. 
Using DL methods allow us to overcome the language pre-defining issue present in current computer simulation methods in evolutionary linguistics as in \cite{steels2005emergence} and \cite{cangelosi2012simulating}.
The issue consists in having all basic linguistic elements (such as symbols and rules of generating phrases) to be pre-specified instead of being invented from scratch.
In contrast to previous works \cite{mordatch2018emergence, cao2018emergent} which focus on the emergence of referential signalling systems, we explore the emergent compositionality of the \textbf{non-referential} concept of numerals (which will be explained in Section~\ref{ssec:input_represents}) by designing a \textbf{referential} game in which agents need to transmit numerical concepts to communicate successfully.

Inspired by \cite{kirby2015compression}, we model the emergence of communication protocols in dyads (i.e. the smallest possible social group of two agents) that are nodes in iterated learning chain \cite{kirby1999function}. 
We observe that iterated learning can facilitate the emergence of compositional languages for numeric concepts.
However, the emergence of languages with such properties depends on the representation of numerical concepts present in the objects observed by the agents during the training.
To be specific, compositional languages emerge when numeric concepts are: i) represented as a concatenation of one-hot vectors directly representing numbers; ii) implied in images of scenes featuring different number of objects.
Further, we show that input representations influence the difficulty of learning a particular language by the agents, which explains the different results in case of iterated learning.
For numerical concepts, we, therefore, argue that one necessary condition for the emergence of compositional languages in iterated learning is that these languages can be fully learnt
\footnote{A language is said to be fully learnt if: i) a speaker can always reproduce same messages as in the language given the inputs; ii) a listener could always obtain 100\% accuracy given only the messages in it.}
with less iterations for agents (especially listeners), compared with holistic languages and emergent languages from dyads.


\section{Model Methods}
\label{sec:2.game_method}

\subsection{The Bag-Select Game}
\label{ssec:2.1.game}

To test whether computational agents can learn to transmit numerical concepts, we propose a referential game called as ``Bag-Select'' game which is briefly illustrated in Figure 1.

\begin{figure}[!h]
  \label{fig:refer_game}
  \centering
  \includegraphics[width=0.6\textwidth]{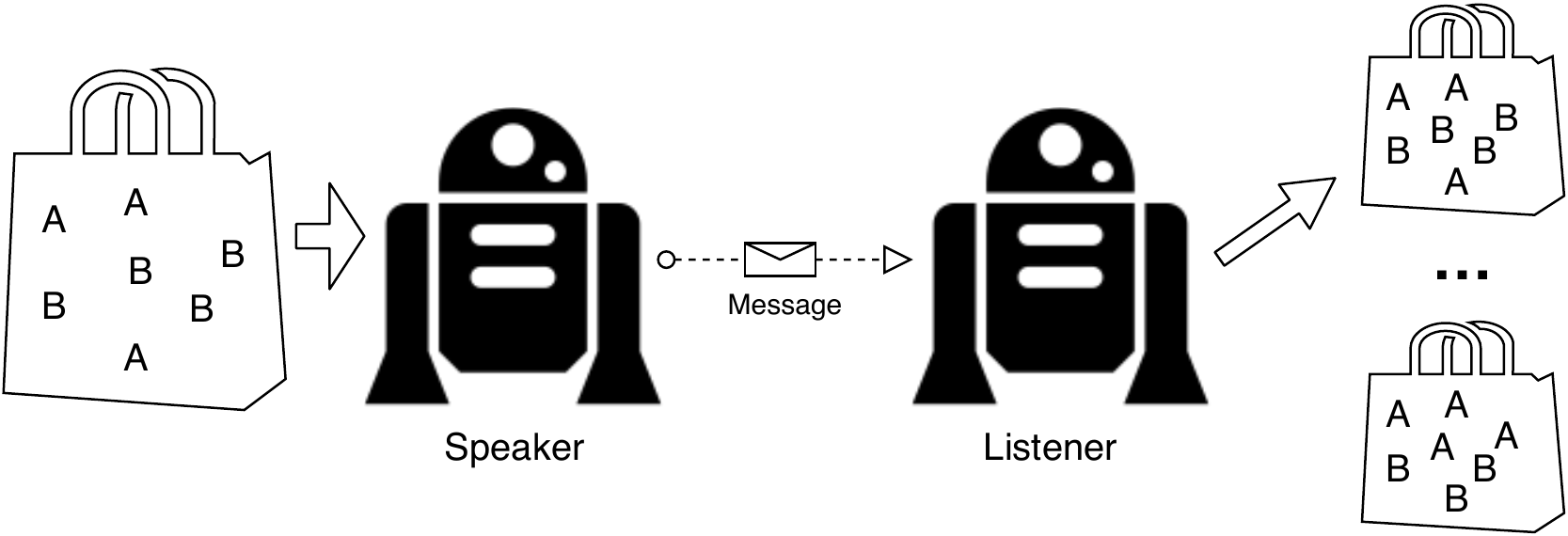}
  \caption{{\footnotesize Sketch diagram of the Bag-Select game.
  The speaker observes a bag of objects of distinct types.
  The bag can contain a different number of objects of the specific type (here, three As and four Bs).
  The speaker produces a message, and the listener uses it to select the bag, that the speaker initially observed.
  The original bag is contained in a set among several other distinct bags, which differ only in the number of As and Bs.}}
\end{figure}

Note that there are always 15 candidates for listeners to choose from in our game. More details about our game setting are given in Section A in Appendix.

\subsection{Input Representations of Bags to be Communicated}
\label{ssec:input_represents}

The overall architecture of our implementation is similar to communication models proposed by \cite{havrylov2017emergence}.
However, unlike theirs, in our game, an input $b_i$ can be

1. \textbf{Concatenation}: a concatenation of one-hot vectors that represent numbers of each kind of objects, e.g. ``2A3B'' (a bag containing 2 As and 3 Bs) would be represented as $[0 0 1 0 0 0; 0 0 0 1 0 0]$ and ``2A0B'' would be represented as $[0 0 1 0 0 0; 1 0 0 0 0 0]$.

\begin{figure}[!h]
  \centering
  \subfigure[``0A0B'']{
  \label{fig_sub:img_inputs1}
  \includegraphics[width=0.18\textwidth]{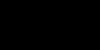}}
  \subfigure[``0A2B'']{
  \label{fig_sub:img_inputs2}
  \includegraphics[width=0.18\textwidth]{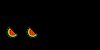}}
  \subfigure[``2A0B'']{
  \label{fig_sub:img_inputs3}
  \includegraphics[width=0.18\textwidth]{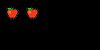}}
  \subfigure[``2A3B'']{
  \label{fig_sub:img_inputs3}
  \includegraphics[width=0.18\textwidth]{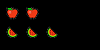}}
  \subfigure[``5A5B'']{
  \label{fig_sub:img_inputs3}
  \includegraphics[width=0.18\textwidth]{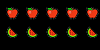}}
  \caption{Example of an image representation of input bags, that contain numerical properties. Captions under each sub-figures indicate the corresponding meaning.}
  \label{fig:img_inputs}
\end{figure}

2. \textbf{Image}: an image containing different numbers of objects, e.g. ``0A0B'', ``0A2B'', ``2A0B'', ``2A3B'', ``5A5B'' would be represented as Figure \ref{fig:img_inputs} (a-e) respectively.

3. \textbf{Bag}: a bag of one-hot vectors that represent the quantity of different types of objects, e.g. ``2A3B'' and ``2A0B'' would be represented as $\{[0 1], [0 1], [1 0], [1 0], [1 0]\}$ and  $\{[0 1], [0 1]\}$ respectively. 

As there is no specific value that can be referred to as numbers of an object in our Image and Bag representations, numeric concepts are \textbf{non-referential} in our games.

Different types of inputs require different encoders, thus we use: i) multilayer perceptron (MLP) for concatenations; ii) the convolutional neural network (CNN) which shares the same architecture of LeNet-5 proposed by \cite{lecun1998gradient} for images; iii) Bag-Encoder for bags.

Our bag-encoder shares almost the same architecture as the set encoder proposed by \cite{vinyals2015order}, except that we replace the softmax function in equation (5) of \cite{vinyals2015order} with the sigmoid function.
Thus, we could keep the feature representation invariant under reordering of the vectors in bags, and avoid introducing normalizing bias (i.e. softmax output has to sum to one) which allows proper encoding of the numbers in the distributed representation of the bag.

To keep both meaning space and message space limited and thus analysable, there are only $2$ different types of objects in our game and the maximum number of each kind of objects is $5$.
Therefore, the size of our Concatenation/Image dataset is 36, and the size of Bag dataset is 35 (excluding the empty bag).
Messages are strings of characters of maximum length 2, where there is an available vocabulary of 10 characters.

\subsection{Iterated Learning for Deep Learning Models}
\label{ssec:2.4.IL}

We contrast two types of the population model.
Following \cite{havrylov2017emergence}, we model dyads, pairs of agents who interact repeatedly and update their network parameters to maximise communicative success.
Following \cite{kirby2015compression}, we contrast the communication systems that emerge in dyads with those that develop in iterated learning transmission chains.
In the latter case, each generation in the chain consists of a pair of agents who are first trained on input-message pairs produced by the previous generation, then update their network parameters during communication with each other to maximise communicative success, before finally generating more data to pass to the next generation.
More details about iterated learning for deep learning models are given in Section B in Appendix.

Besides, the metrics and evaluation methods used in the following experiments are illustrated in Section C in Appendix.


\section{Emergence of Compositional Languages}
\label{sec:experiment_analysis}

In this section, we show that compositional languages can emerge under iterated learning, but only for the Concatenation and Image representations.
As training iterated learning on deep learning models is extremely time-consuming, we report results for only one run per condition.
During the exploratory phases of our research, we conducted multiple runs and found that the variance of resulting patterns of emergent languages is small, which gives us confidence that these results are representative.

\begin{figure}[!h]
  \centering
  \subfigure[Concatenation]{
    \label{fig_sub:cancat_compare}
    \includegraphics[width=0.3\textwidth]{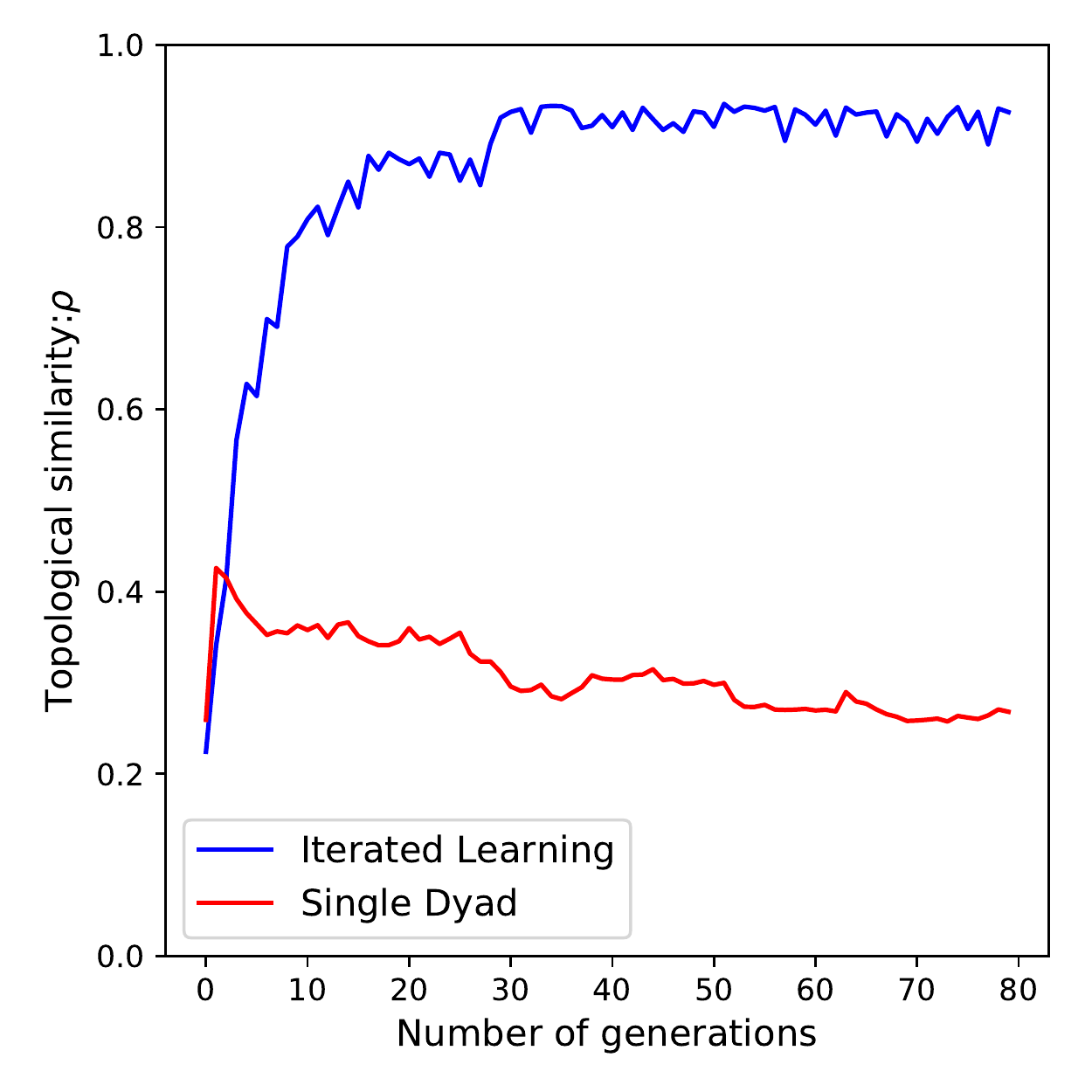}
  }
  \subfigure[Image]{
    \label{fig_sub:img_compare}
    \includegraphics[width=0.3\textwidth]{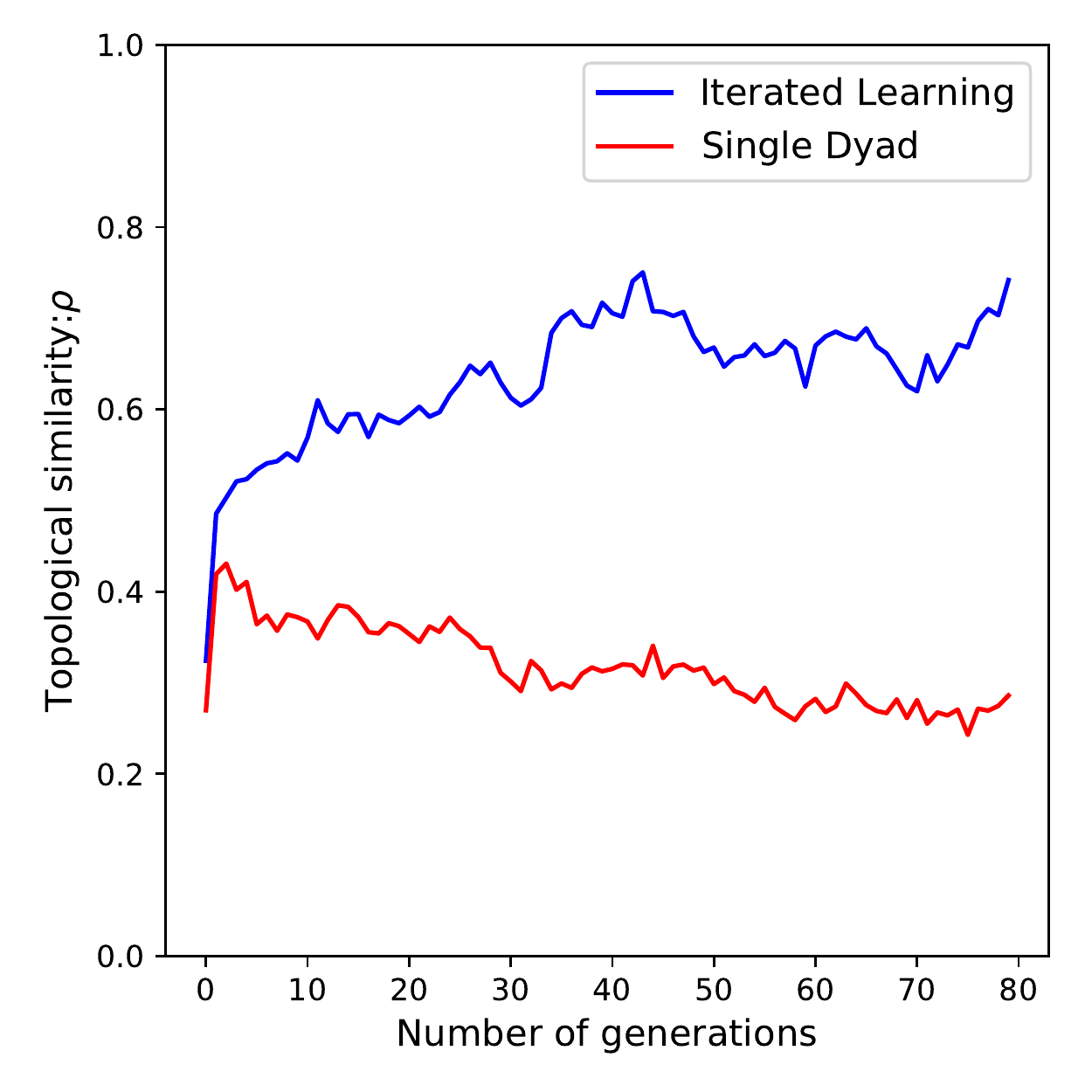}
  }
  \subfigure[Bag]{
    \label{fig_sub:bag_compare}
    \includegraphics[width=0.3\textwidth]{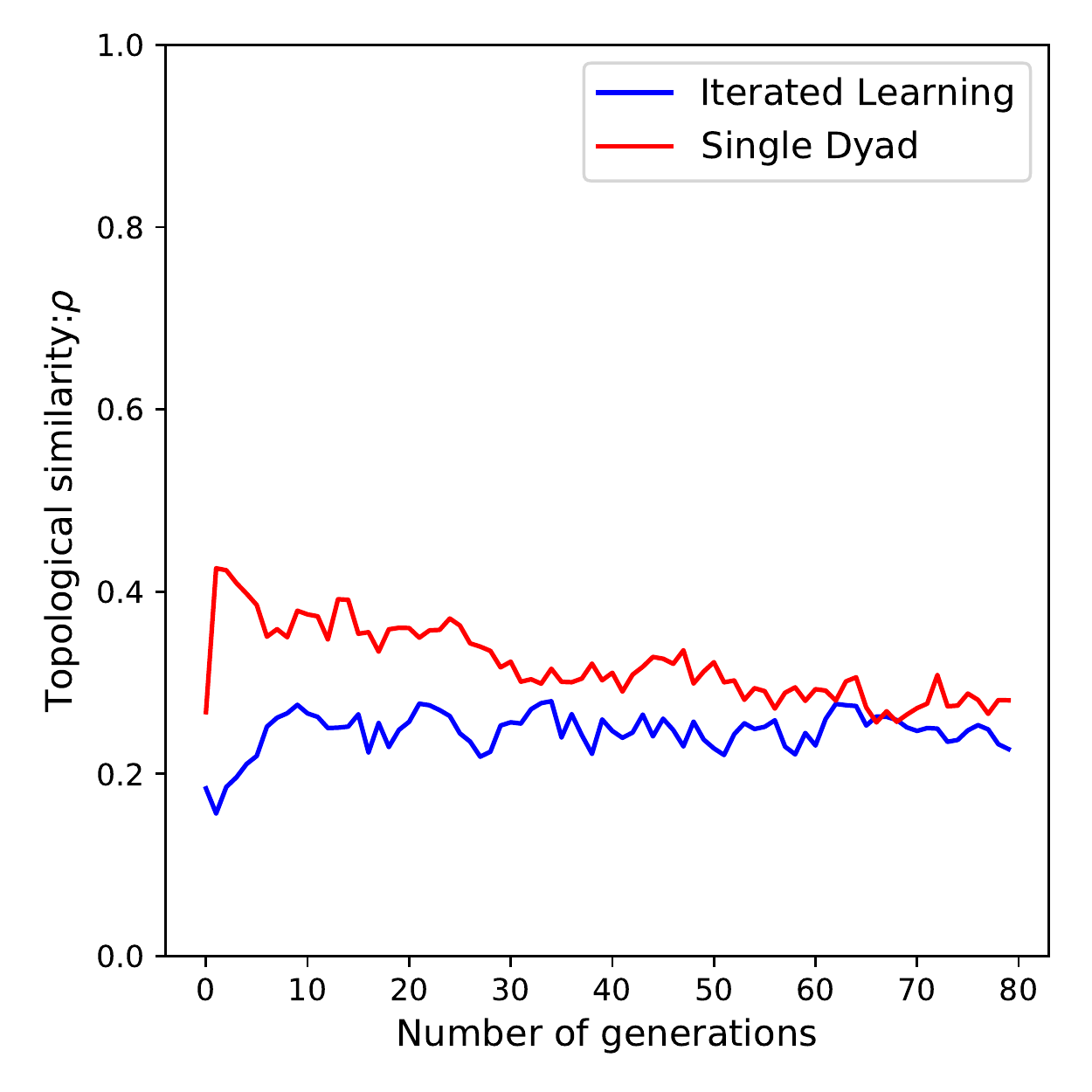}
  }
  \caption{Topological similarity changes over generations on different input representations.}
  \label{fig:effectiveness}
\end{figure}

To verify that iterated learning could successfully amplify the probability density of languages having high compositionality, we track the change of topological similarity of languages having greatest probability density over generations.
The results for the Concatenation, Image and Bag input representations are shown in Figure~\ref{fig:effectiveness}.
As can be seen from the graphs, dyads do not converge on compositional languages under any input representation.
However, in iterated learning models, the topological similarity of emergent languages keep increasing on Concatenation and Image representation.
We also track the how posterior probability of languages change over generations, and the results as well as corresponding final emergent languages are given in Figure \ref{app_fig:effectiveness} in Appendix.

\section{Learnability of Compositional and Emergent Languages }
\label{sec:ex_learnability}

According to \cite{kirby2015compression}, the structure of natural languages is a trade-off between expressivity that arises during communication and compressibility that arises during learning.
Meanwhile, \cite{li2019ease} propose a hypothesis that compositional languages should be easier for listeners to learn than other less structured languages.
Inspired by both of them, we hypothesise that the different effectiveness of iterated learning for different input representations observed in the above experiments is caused by different learnability of compositional languages for different input representations.


To test this hypothesis, we examine the learnability of three language types (compositional, emergent, holistic) for speakers and listeners. The establishment of these different types of languages are illustrated in Section E in Appendix. Meanwhile, training listeners separately is also illustrated there.

\begin{figure}[!h]
  \centering
  \subfigure[On Concatenation]{
    \makebox[0pt][r]{\makebox[10pt]{\raisebox{40pt}{\rotatebox[origin=c]{90}{\scriptsize Listener learning curves}}}}
    \label{fig_sub:cancat_lan_learn}
    \includegraphics[width=0.25\textwidth]{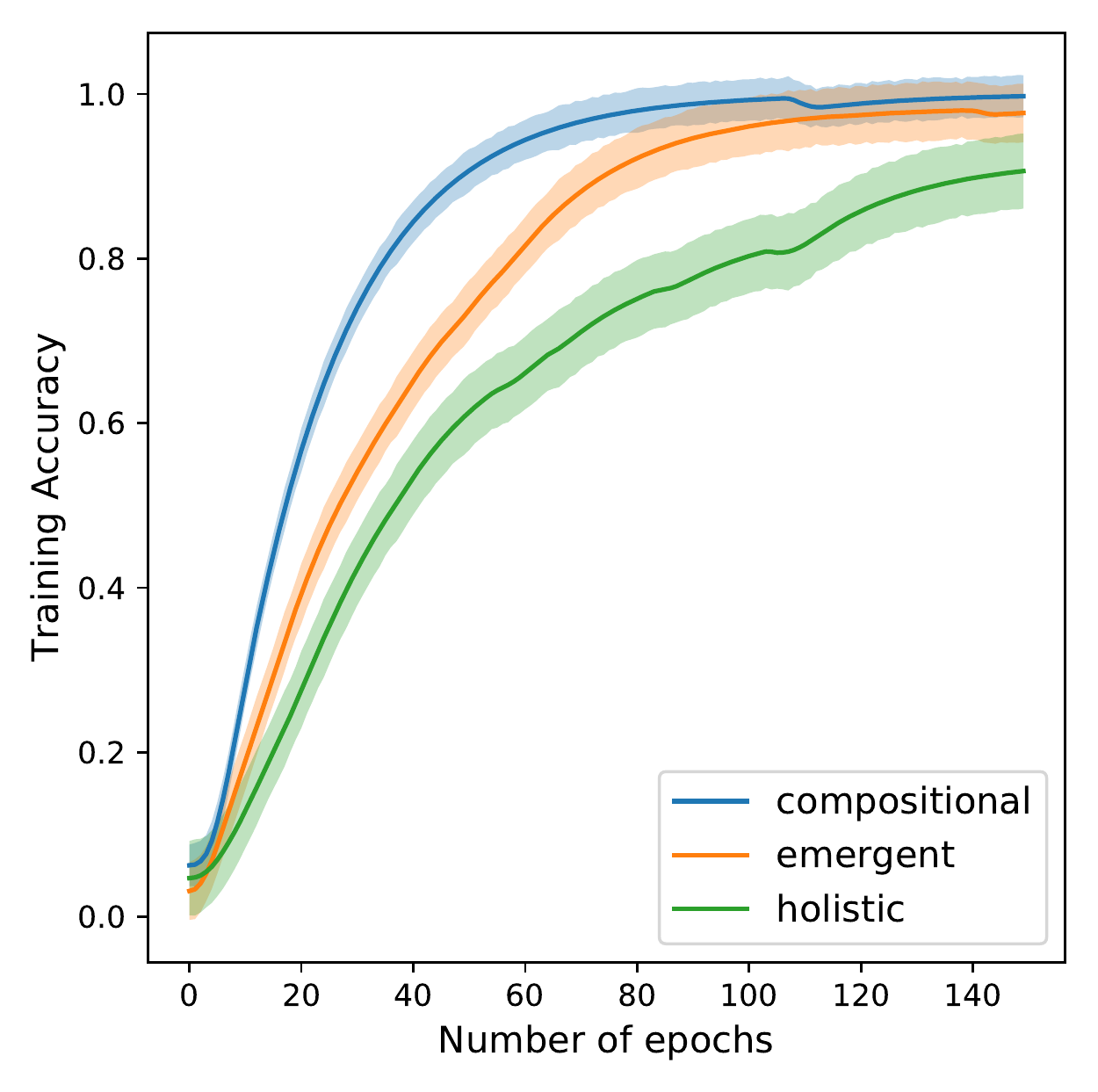}
  }
  \subfigure[On Bag]{
    \label{fig_sub:bag_lis_learn}
    \includegraphics[width=0.25\textwidth]{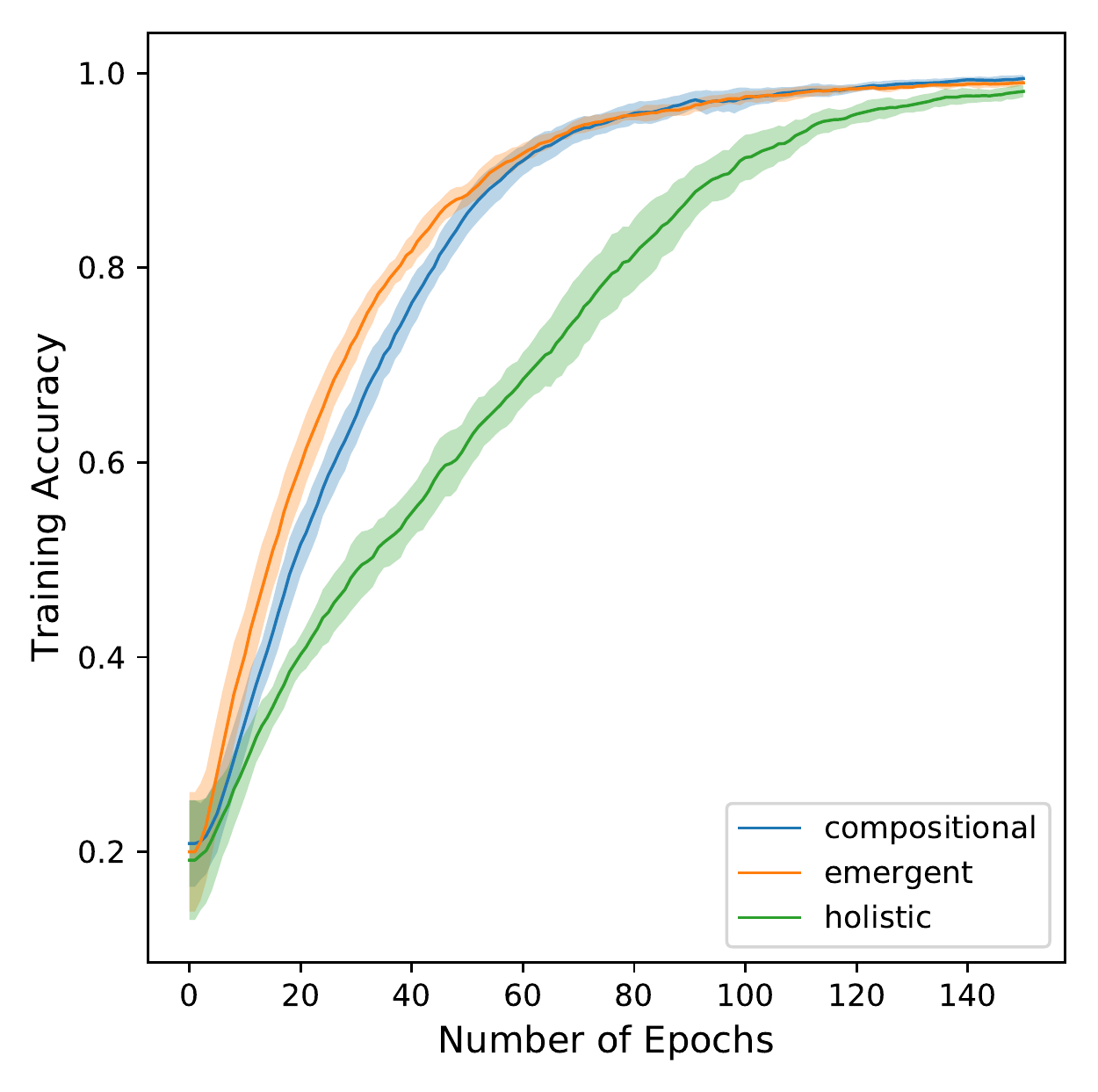}
  }
  \subfigure[On Image]{
    \label{fig_sub:img_lis_learn}
    \includegraphics[width=0.25\textwidth]{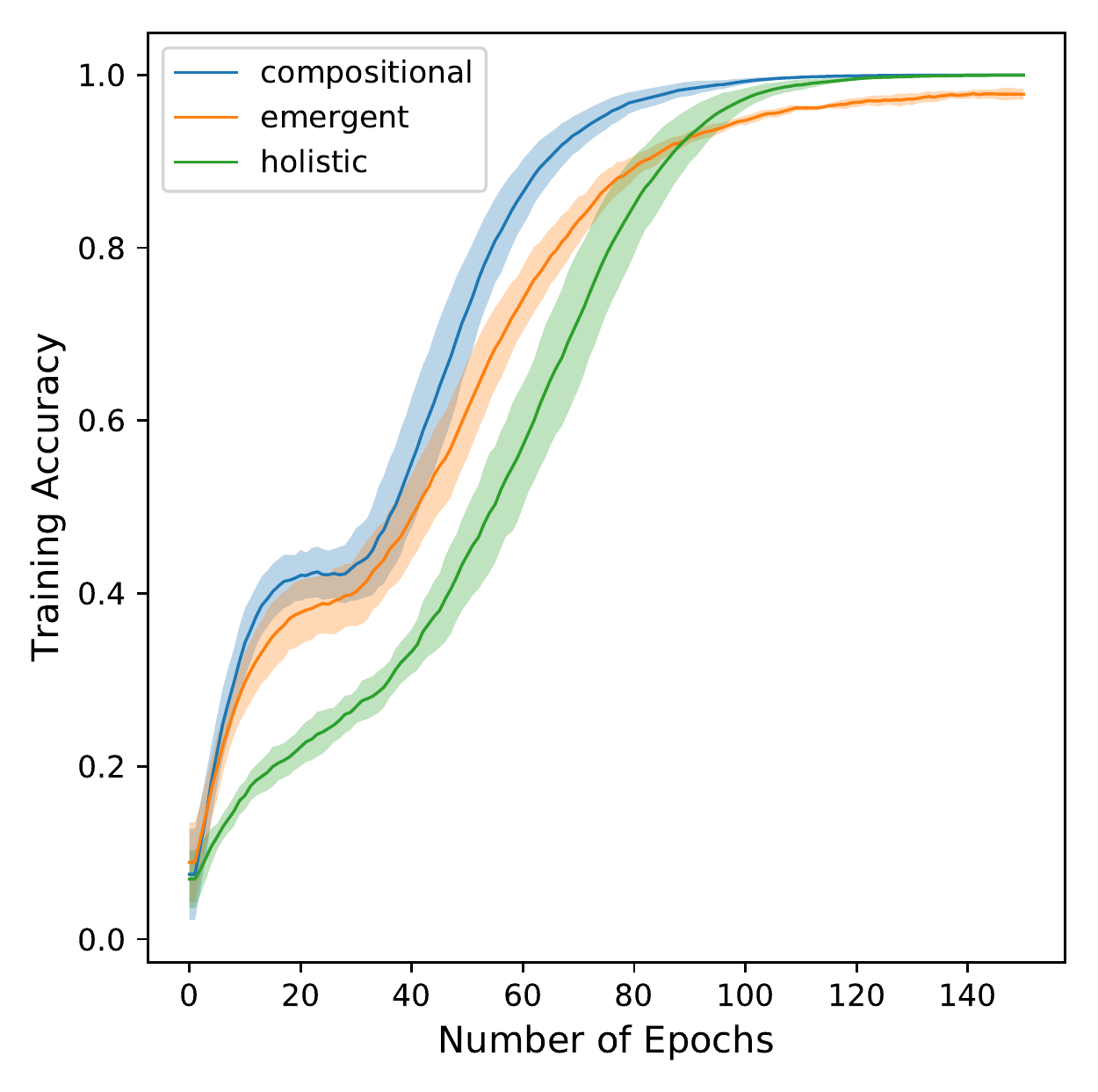}
  }
  \subfigure[On Concatenation]{
    \makebox[0pt][r]{\makebox[10pt]{\raisebox{40pt}{\rotatebox[origin=c]{90}{\scriptsize Speaker learning curves}}}}
    \label{fig_sub:concat_spk_learn}
    \includegraphics[width=0.25\textwidth]{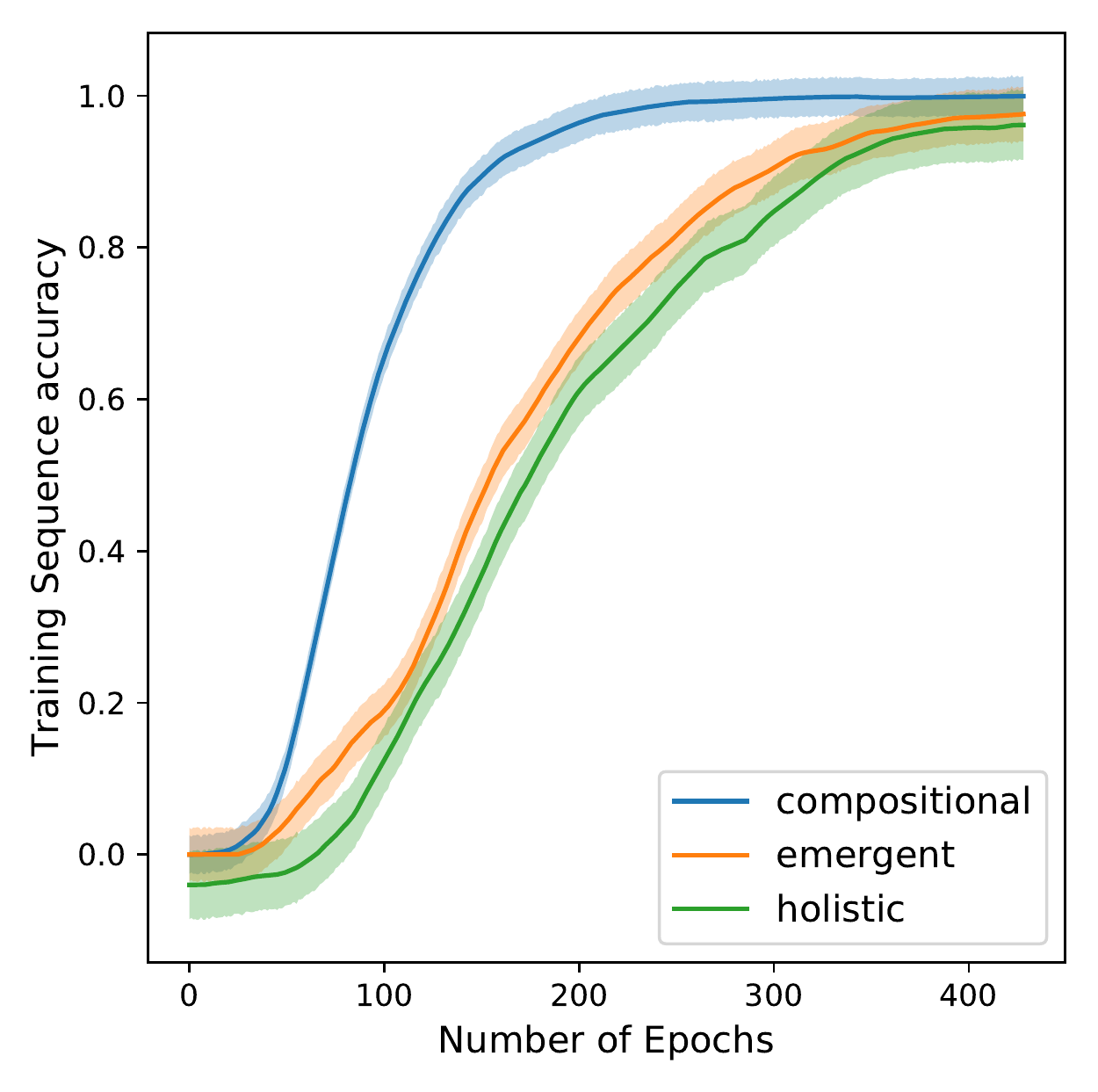}
  }
  \subfigure[On Bag]{
    \label{fig_sub:bag_spk_learn}
    \includegraphics[width=0.25\textwidth]{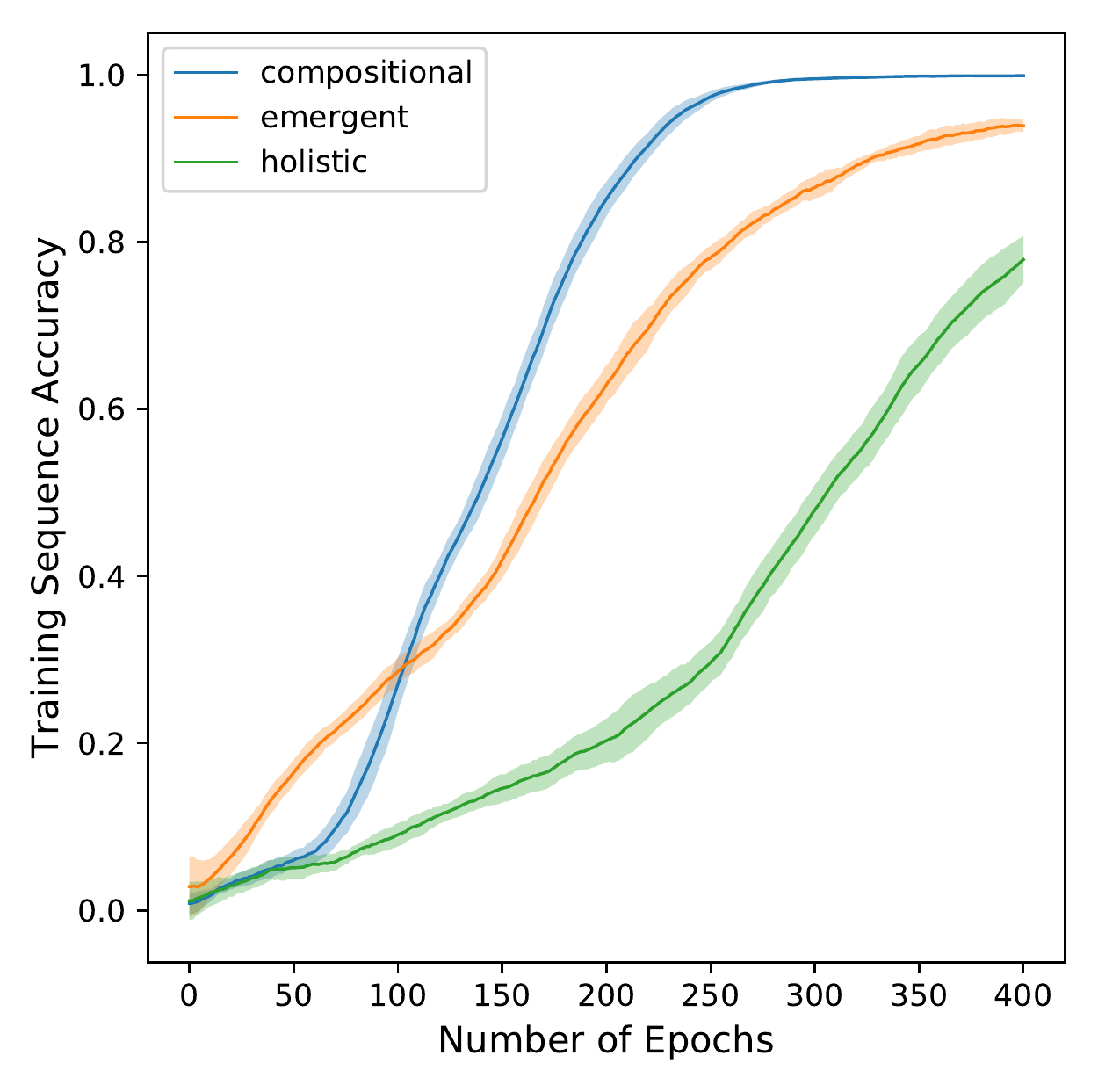}
  }
  \subfigure[On Image]{
    \label{fig_sub:img_spk_learn}
    \includegraphics[width=0.25\textwidth]{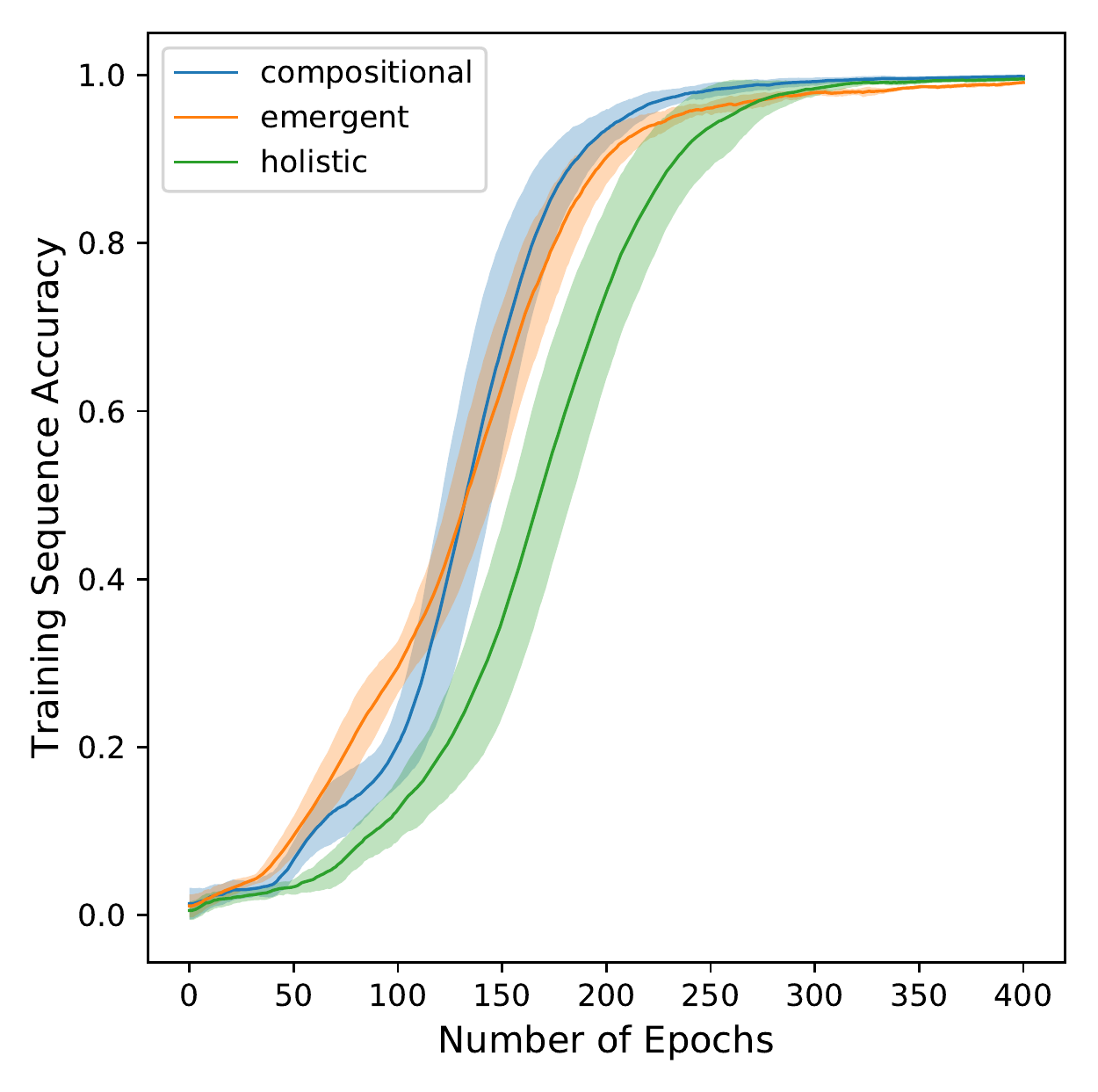}
  }
  \caption{Experiments results on learnability of different kinds of languages, the first row is for listeners and the second row is for speakers. input representations are given below each sub-figure. The lines are means of 10 runs with different random seeds, and the corresponding standard deviations are shown by the shadow area around the lines.}
  \label{fig:learnability}
\end{figure}

The learning curves of both listeners and speakers on different input representations are shown in Figure~\ref{fig:learnability}.

It is clear from Figure~\ref{fig:learnability} that compositional languages require fewer training iterations than the other 2 kinds of languages in almost all the cases, with two exceptions: i) emergent languages has better learnability for listeners on the Bag representation; ii) compositional and emergent languages have almost the same learnability for speakers on the Image representation. Further explainations are provided in Section F in Appendix.

\section{Conclusion}
\label{sec:conclusion}

We use the Bag-Select game to demonstrate that iterated learning leads to the emergence of compositional languages for transmitting numeric concepts.
However, this result is dependent on the representations of inputs, and its effectiveness depends on that compositional languages have the optimal learnability for listeners in the communication game.
While our findings confirm that structure of languages emerges under the pressure of both expressivity and learnability, at least for deep learning agents, the representation of the input representations affects on learnability and therefore on the structure of the emergent languages.

%

\clearpage

\bibliographystyle{plain}
\bibliography{evolang12}

\clearpage

\section*{Appendix}

\subsection*{A: Game Description}
\label{app:game_desciption}

In our game settings, there are two different kinds of agents: i) speaker $S$ that  observes the input $b_i$ at the beginning of round $i$ and then generates a message $m_i$; ii) listener $L$ that receives $m_i$ and then selects $\hat{b}_i$ among candidates $c_i^k$ where $k\in \{1, 2, \dots, 15\}$.
In our experiments, there are always 15 candidates, among which one would be $b_i$ and the other fourteen would be uniformly sampled from the whole meaning space excluding $b_i$, for listeners to choose from.
The game only succeeds if $\hat{b}_i$ matches $b_i$.
The speaker does not have access to the entire candidate list, only to the correct bag $b_i$, which implies that the number of each object type has to be encoded in the message in order to reliably succeed in the game.

\subsection*{B: Phases of Iterated Learning for Deep Learning}

To be more specific, each generation in our iterated learning model includes the following three steps:

\begin{enumerate}
  \item \textbf{Learning phase}: During this phase, we train speaker $S_t$ separately to reproduce same messages given the inputs, with the input-message pairs generated by $S_{t-1}$.
  For example, an input-message pair is $\mbox{``1A0B''} \rightarrow \mbox{``yw''}$, then we would train speakers to produce ``yw'' given the input ``1A0B''.
  To do so, we use stochastic gradient descent (SGD) \cite{robbins1951stochastic} to update parameters of $S_t$. Gradients are computed using the back-propagation \cite{rumelhart1988learning} algorithm with the cross entropy loss function between speaker's predictions and the messages generated by $S_{t-1}$.
  The number of training iterations is fixed such that predefined compositional language can be fully learnt (note that language produced by $S_{t-1}$ is not necessarily compositional). 
  There is no such phase in the first generation of iterated learning chain, as there are no input-message pairs for training $S_1$.
  \item \textbf{Interaction phase}: During this phase, we train $S_t$ and $L_t$ agents to play the communication game using SGD.
  The reward is represented by the negative cross entropy between the probability distribution of the listener's prediction and the one-hot representation of the correct bag.
  Analogous to linguistic symbols, i.e. words, the messages transmitted between dyad should contain only discrete symbols.
  However, discrete messages would make learning prohibitively expensive from the computational perspective for computing the gradients would require enumeration of all possible messages.
  To overcome this limitation, we use the Gumbel-softmax estimator proposed by \cite{jang2016categorical} to train our models.
  Besides, we set the number of iterations here to be fixed over generations, and number of iterations is obtained by pre-training a dyad to promise that it is long enough for a dyad to obtain $100\%$ communication success rate.
  \item \textbf{Transmission phase}: During this phase, we feed all $b_i$ in the training set into $S_t$ and sample messages $m_i$ based on the generated probability distribution over vocabulary.
  This builds a dataset of input-message pairs for $S_{t+1}$ to learn from.
  In addition, the number of sampled input-message pairs is $2,000$ so that they effectively reflect the distribution of all possible languages - note that since there are only 35-36 distinct input meanings to be communicated, there is no data bottleneck here, and learners will see signals for the entire space of possible meanings.
\end{enumerate}

Additionally, interaction phase is the same as training dyad models like \cite{havrylov2017emergence}.

\subsection*{C: Metrics and Evaluations}
\label{app_sec:metric_eval}

Following \cite{brighton2006understanding}, we take the topological similarity between meaning space and message space as the metric for measuring compositionality of languages, and we use Hamming distance and edit distance with respect to meaning space and message space.
Equivalently, the topological similarity becomes the correlation coefficient between the Hamming distances between pairs of meanings and the edit distances between their corresponding messages.
This measure captures the intuition that, in a compositional language, similar meanings will be conveyed using similar signals.
We denote this measure of topological similarity as $\rho$; holistic (non-compositional) languages will have $\rho$ scores around 0, a perfectly compositional language will have a $\rho$ score of close to 1. 

Additionally, we also need to measure the learning performance of new learners in order to compare the learnability of different languages, which is illustrated in Section~\ref{sec:ex_learnability}.
To do so, we use the accuracy of reproducing messages (both sequence-level and token-level) for speakers and accuracy of choosing the correct candidate for listeners respectively.

\subsection*{D: Experiments on the Emergence of Compositional Languages}

\begin{figure}[!h]
  \centering
  \subfigure[]{
    \makebox[0pt][r]{\makebox[10pt]{\raisebox{50pt}{\rotatebox[origin=c]{90}{\scriptsize Concatenation input representation}}}}
    \label{fig_sub:cancat_compare}
    \includegraphics[width=0.3\textwidth]{graph/concat_sim_compare.pdf}
  }
  \subfigure[]{
    \label{fig_sub:cancat_lan_p_change}
    \includegraphics[width=0.3\textwidth]{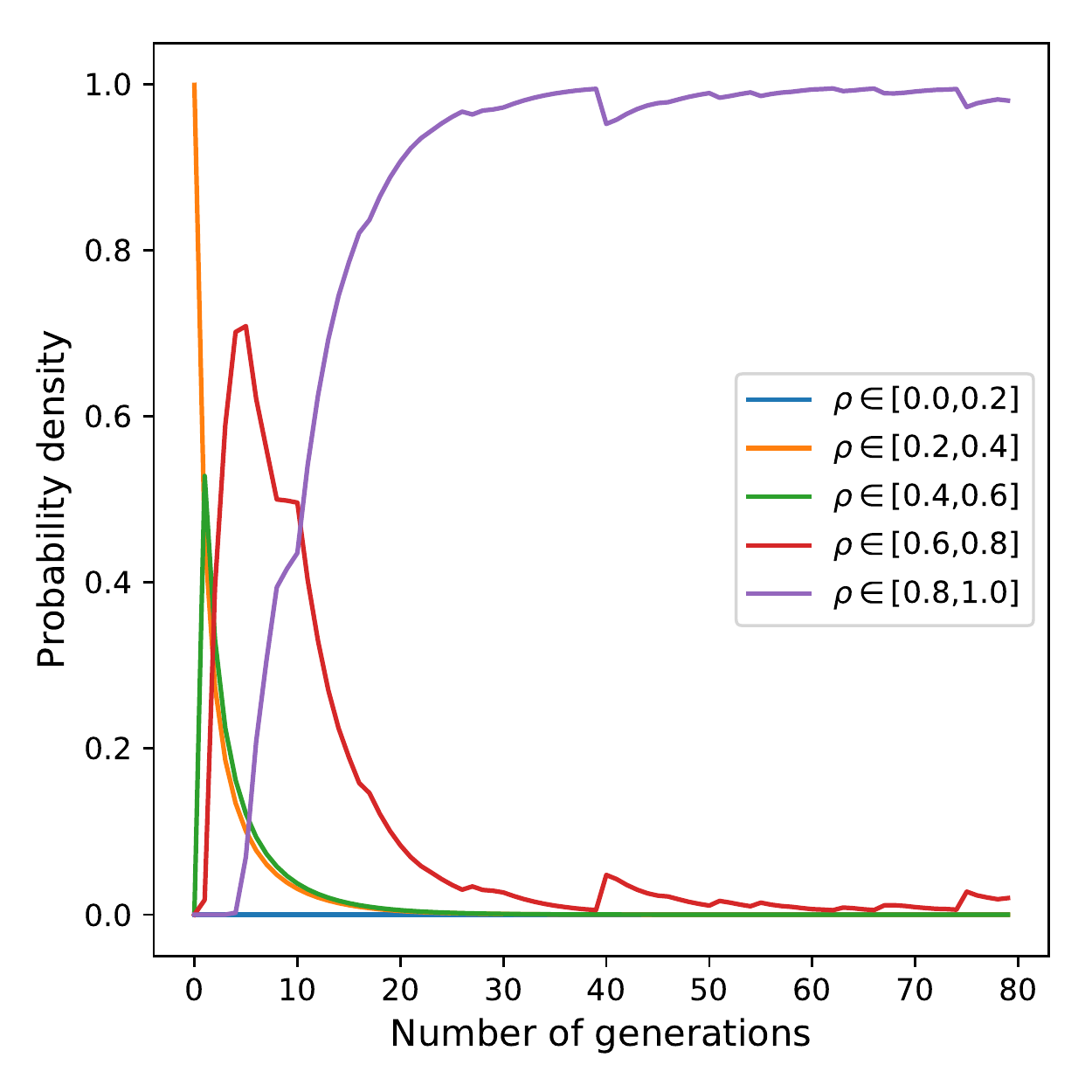}
  }
  \subfigure[]{
    \label{fig_sub:cancat_emergetn_lan}
    \includegraphics[width=0.3\textwidth]{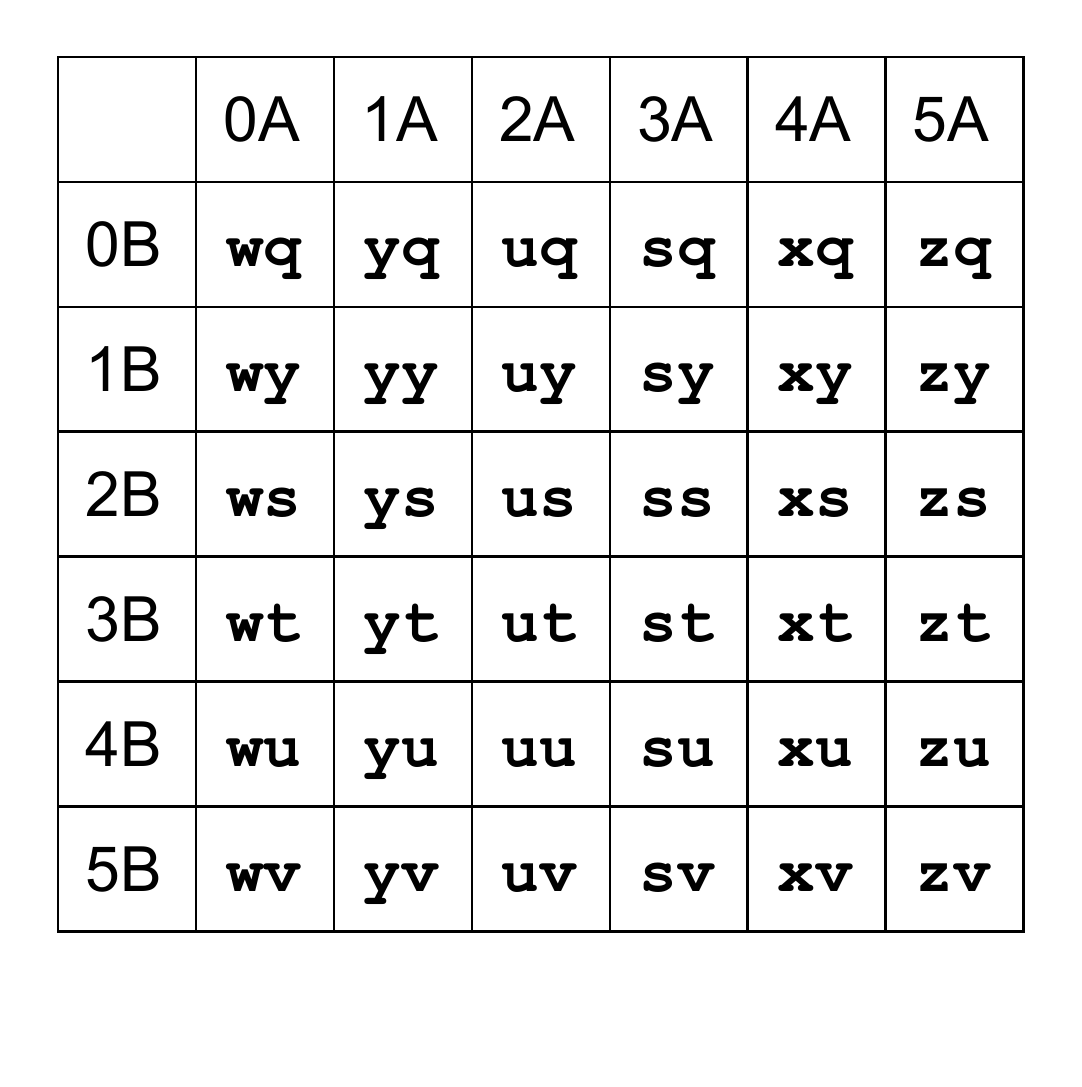}
  }
  \subfigure[]{
    \makebox[0pt][r]{\makebox[10pt]{\raisebox{50pt}{\rotatebox[origin=c]{90}{\scriptsize Image input representation}}}}
    \label{fig_sub:img_compare}
    \includegraphics[width=0.3\textwidth]{graph/img_sim_compare.pdf}
  }
  \subfigure[]{
    \label{fig_sub:img_lan_p_change}
    \includegraphics[width=0.3\textwidth]{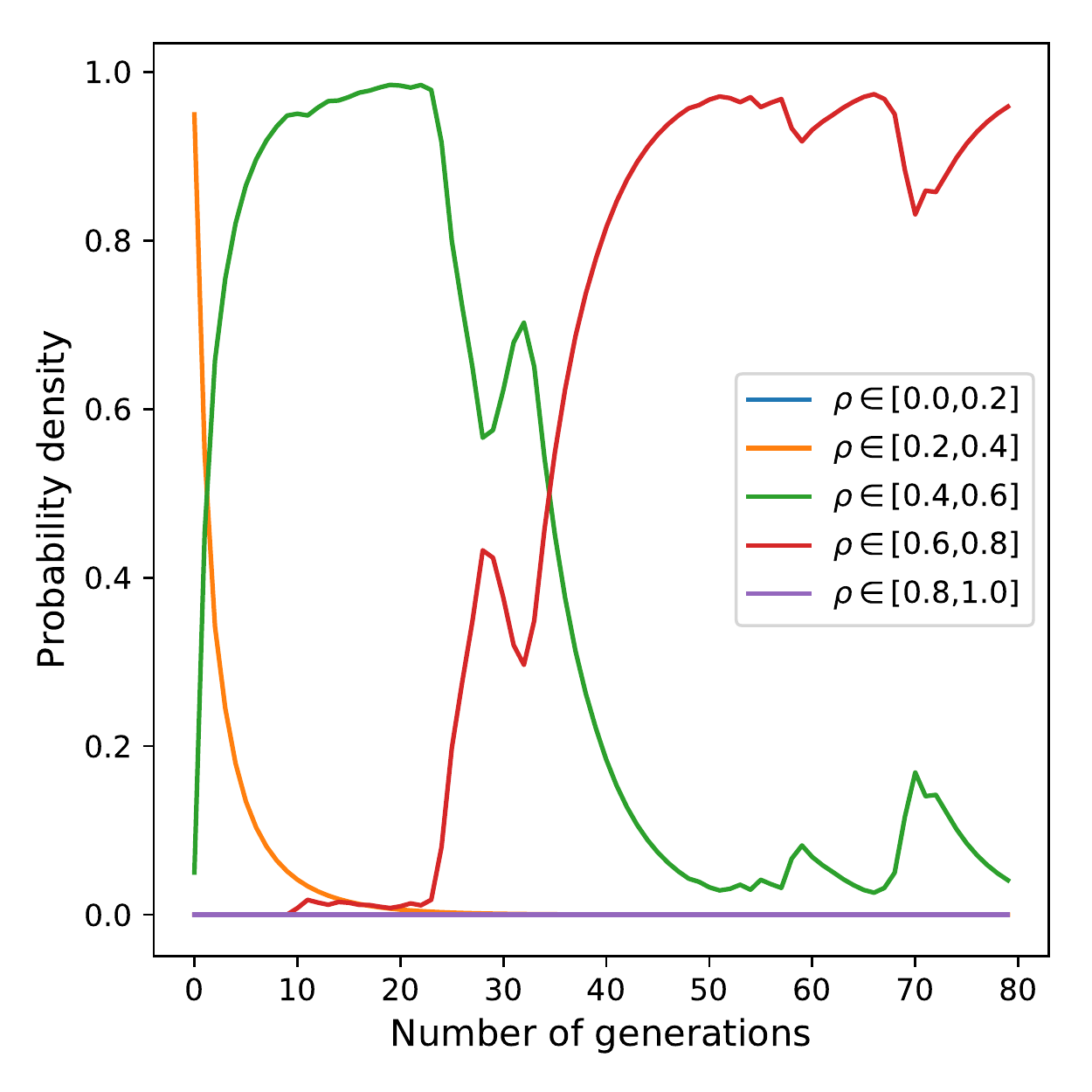}
  }
  \subfigure[]{
    \label{fig_sub:img_emergent_lan}
    \includegraphics[width=0.3\textwidth]{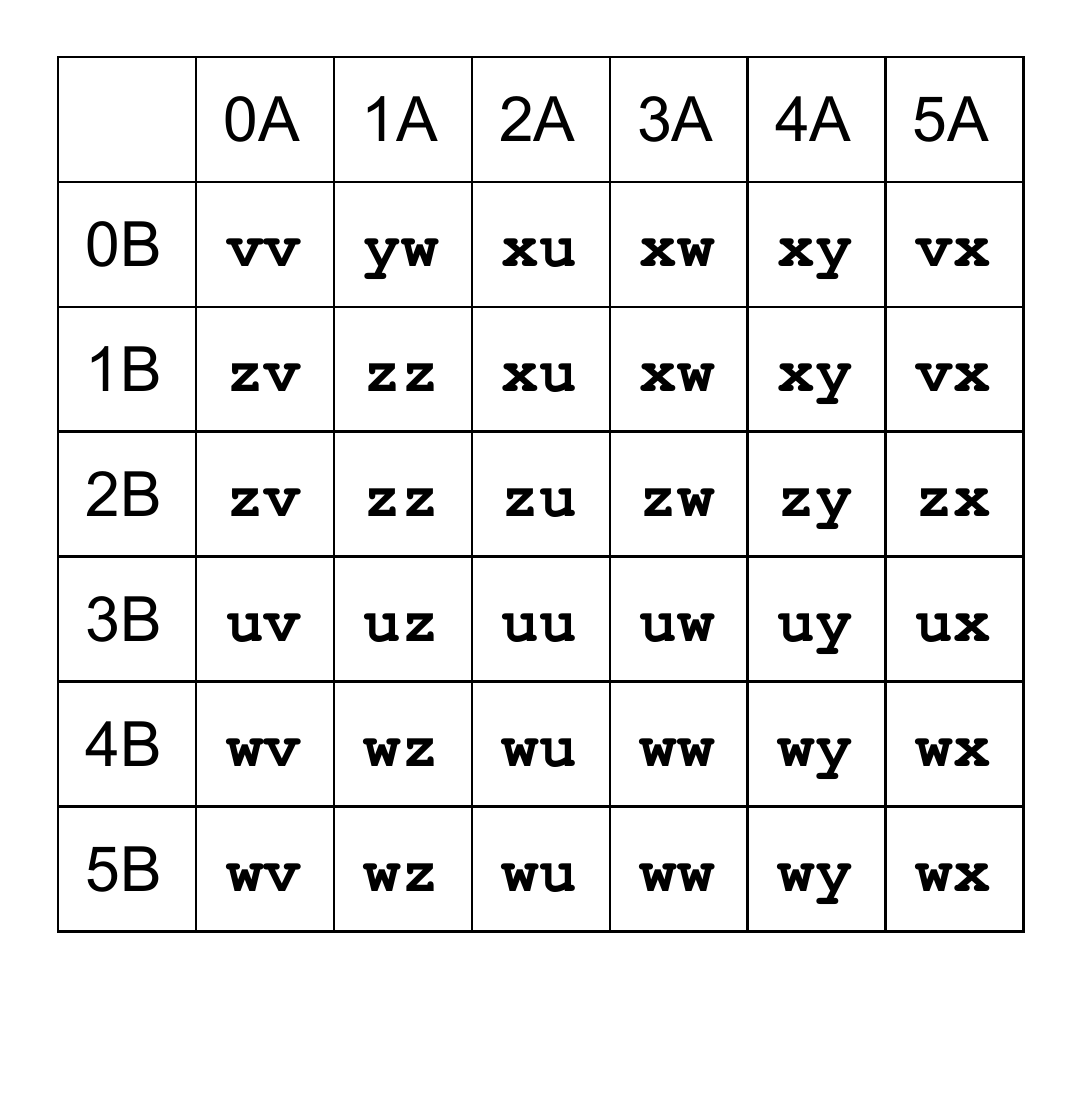}
  }
  \subfigure[]{
    \makebox[0pt][r]{\makebox[10pt]{\raisebox{50pt}{\rotatebox[origin=c]{90}{\scriptsize Bag input representation}}}}
    \label{fig_sub:bag_compare}
    \includegraphics[width=0.3\textwidth]{graph/set_sim_compare.pdf}
  }
  \subfigure[]{
    \label{fig_sub:bag_lan_p_change}
    \includegraphics[width=0.3\textwidth]{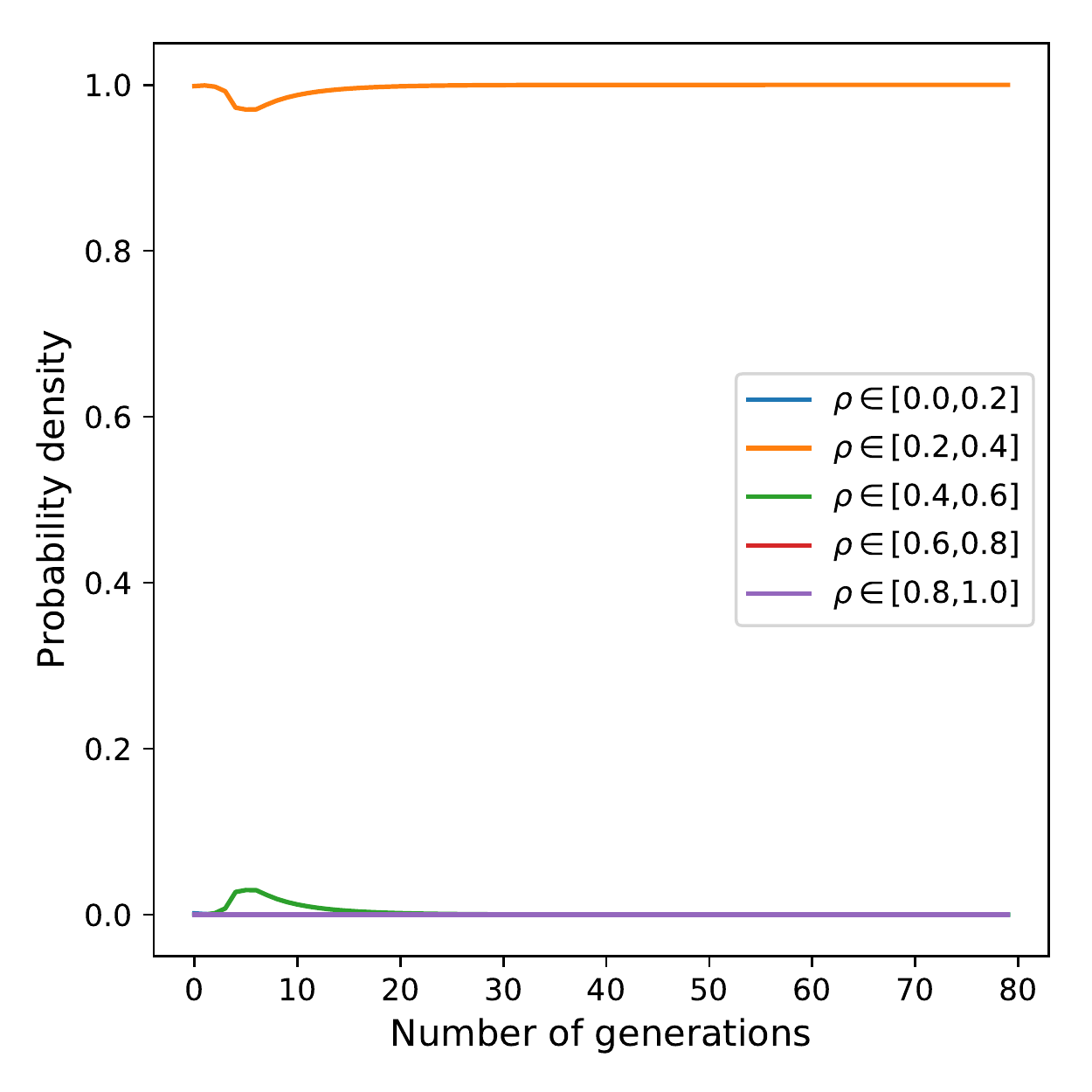}
  }
  \subfigure[]{
    \label{fig_sub:bag_emergetn_lan}
    \includegraphics[width=0.3\textwidth]{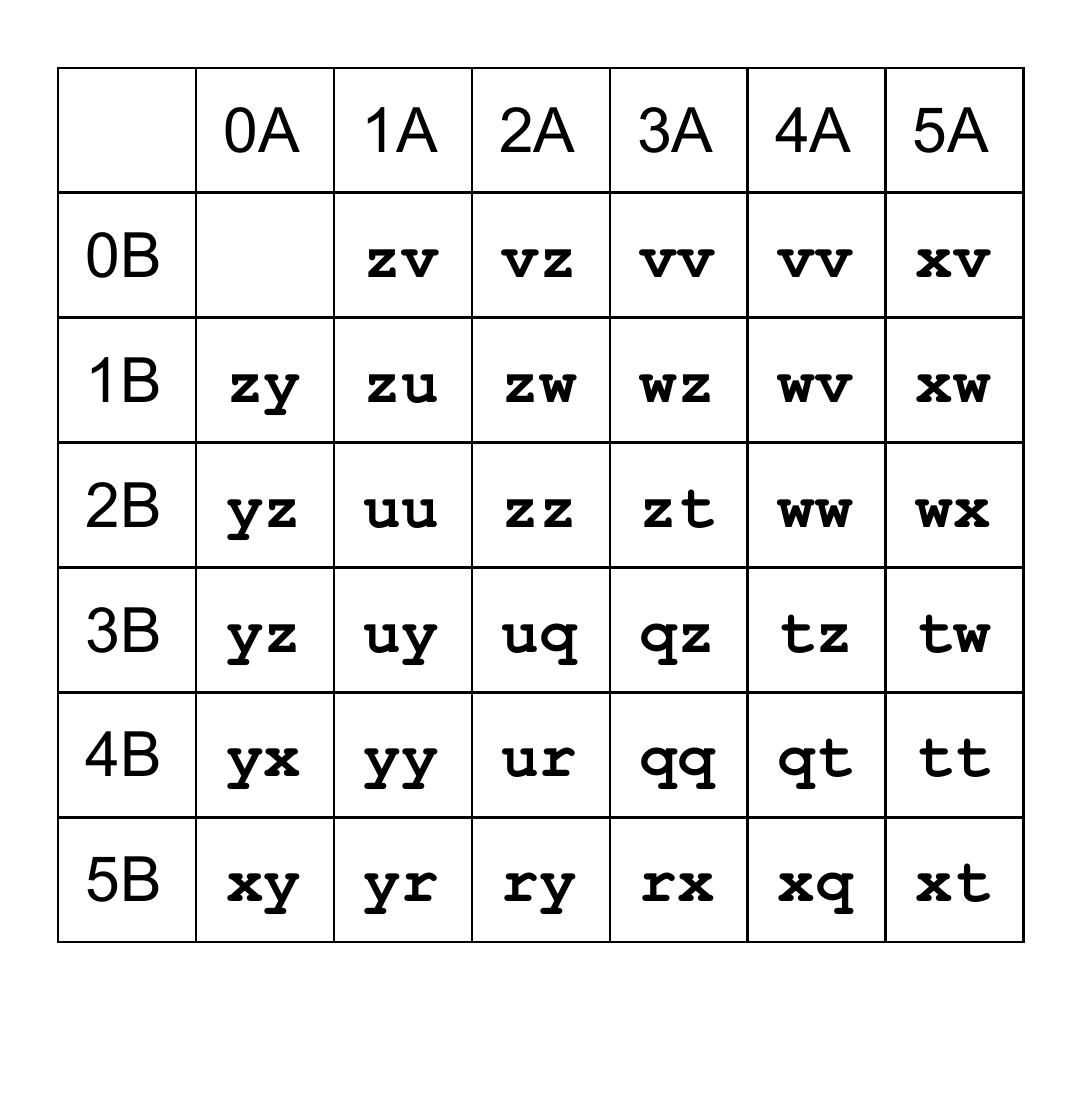}
  }
  \caption{Experiments results on different input representations. The rows from top to bottom are results for Concatenation, Image and Bag representations respectively. The columns from left to right are: i) smoothed topological similarity (of language having greatest probability)  over generations with different population models; ii) smoothed posterior probability of languages having different compositionality ($\rho$) over generations; iii) final emergent language facilitated by iterated learning, where the first row and first column are numbers of object ``A'' and ``B'' respectively.}
  \label{app_fig:effectiveness}
\end{figure}

In our experiments illustrated in Section \ref{sec:experiment_analysis}, we tracked: i) changes of topological similarity  of emergent languages over generations; ii) changes of posterior probability of languages having different compositionality over generations. The results are shown in Figure \ref{app_fig:effectiveness} as follow, which also includes the final emergent languages (i.e. the languages having greatest probability density after training).

As can be seen from the above figure, in iterated learning models, the probability of languages with high compositionality ($\rho > 0.6$) keeps increasing over generations and gradually dominates all other languages, for the Concatenation and Image input representations; compositional languages do not develop in the Bag input representation.
The compositional structure in the languages that emerge under the Concatenation input is clear from the example language given in Figure \ref{app_fig:effectiveness} (rightmost column), as is the absence of compositionality in the example language that develops under the Bag encoding; the final emergent language on Image representation is not perfectly compositional but contains a high degree of regularity.

\subsection*{E: Establish and Train Different Types of Languages}

Our compositional test language was hand-designed and resembled the compositional languages that emerge under iterated learning in the Concatenation condition.
Our holistic language was generated by randomly mapping messages from compositional languages (so that it shares same expressivity as compositional language) to inputs that constitute the whole meaning space.
Our emergent test languages came from a dyad, trained to communicate as per the dyad models, once that dyad obtained 100\% performance -- as such, we would expect them to be largely holistic. 

With these languages, we train speakers separately, which is illustrated in Section B in Appendix.
At the same time, we train listeners separately to correctly complete the game with only messages in a language.
For example, an input-message pair in a language is $\mbox{``1A0B''} \rightarrow \mbox{``yw''}$, then we would train listeners to select ``1A0B'' among the 15 candidates after taking ``yw'' as input.
To do so, we still take the cross entropy between the correct candidate and listener's predicted probability distribution as the loss and apply SGD to update the parameters of listeners.

\subsection*{F: Further Explaination about Learnability Experiments}

Based on the results shown in Figure~\ref{fig:learnability}, considering that the topological similarity of final emergent languages given the Bag representation is much lower than Concatenation/Image representations, we argue that iterated learning will amplify the probability of compositional languages only if less training iterations are necessary for \textbf{listeners} to learn the compositional languages.
As we can theoretically prove that compositional languages always have lower sample complexity than any other non-degenerate languages and thus better learnability for speakers (based on statistical learning theory \cite{vapnik1998statistical}), we actually only need to care about learnability for listeners here, instead of both speakers and listeners as before.
Otherwise, iterated learning does not show lead to an increase in compositionality.
Moreover, our results could also support the hypothesis that compositionality (which is an aspect of linguistic structure) emerges under the pressure of both expressivity and learnability \cite{smith2013linguistic}, considering that emergent languages have better learnability on Bag representation than compositional languages; as such, those languages still represent a trade-off between learnability and expressivity, but under a slightly different learnability constraint.
We are currently investigating why the Bag input encoding makes non-compositional languages more learnable. 

\end{document}